\documentclass[10pt,twocolumn,letterpaper]{article}

\usepackage{cvpr}
\usepackage{times}
\usepackage{epsfig}
\usepackage{graphicx}
\usepackage{amsmath}
\usepackage{amssymb}
\usepackage{multirow}
\usepackage{subfigure}
\usepackage{algorithmic}
\usepackage{amsthm}
\usepackage{booktabs}
\newtheorem{definition}{Definition}
\usepackage[ruled,lined]{algorithm2e}
\usepackage{url}
\usepackage[bookmarks=false]{hyperref}

\cvprfinalcopy % *** Uncomment this line for the final submission

\def\cvprPaperID{1425} % *** Enter the CVPR Paper ID here

\ifcvprfinal\fi
\begin{document}

%%%%%%%%% TITLE
\title{Central Similarity Quantization for Efficient Image and Video Retrieval}

\author{Li Yuan\textsuperscript{\rm 1} \, Tao Wang\textsuperscript{\rm 1}\, Xiaopeng Zhang\textsuperscript{\rm 3} \, Francis EH Tay\textsuperscript{\rm 1}\, Zequn Jie\textsuperscript{\rm 2} \, Wei Liu\textsuperscript{\rm 2}\, Jiashi Feng\textsuperscript{\rm 1}\\\\
\textsuperscript{\rm 1}National University of Singapore \quad \textsuperscript{\rm 2}Tencent AI Lab\quad \textsuperscript{\rm 3}Huawei Noah's Ark Lab\\
{\{ylustcnus, twangnh, zequn.nus\}@gmail.com},\, {wl2223@columbia.edu},\, {\{mpetayeh,elefjia\}@nus.edu.sg}
}
%{zhangxiaopeng12@huawei.com,}\,

\maketitle
\thispagestyle{empty}

%%%%%%%%% ABSTRACT
\begin{abstract}
Existing data-dependent hashing methods usually learn hash functions from pairwise or triplet data relationships, which only capture the data similarity locally, and often suffer from low learning efficiency and low collision rate. In this work, we propose a new \emph{global} similarity metric, termed as \emph{central similarity}, with which the hash codes of similar data pairs are encouraged to approach a common center and those for dissimilar pairs to converge to different centers, to improve hash learning efficiency and retrieval accuracy. We principally formulate the computation of the proposed central similarity metric by introducing a new concept, i.e., \emph{hash center} that refers to a set of data points scattered in the Hamming space with a sufficient mutual distance between each other.
We then  provide an efficient method to construct well separated  hash centers by leveraging the Hadamard matrix and Bernoulli distributions. 
Finally, we propose the Central Similarity Quantization (CSQ) that optimizes the central similarity between data points w.r.t.\ their hash centers instead of optimizing the local similarity. CSQ is generic and applicable to both image and video hashing scenarios. Extensive experiments on large-scale image and video retrieval tasks demonstrate that CSQ can generate cohesive hash codes for similar data pairs and dispersed hash codes for dissimilar pairs, achieving a noticeable boost in retrieval performance, i.e. 3\%-20\% in mAP over the previous state-of-the-arts \footnote{The code is at: \url{https://github.com/yuanli2333/Hadamard-Matrix-for-hashing}}
\end{abstract}.
\vspace{-4mm}
%%%%%%%%% BODY TEXT
\section{Introduction}

\iffalse
\begin{figure}[t!]
\begin{center}
\setlength{\tabcolsep}{1.5pt}
\includegraphics[scale=0.07]{images/motivation.jpg}
%\vspace{0.1cm}
\caption{Comparison of traditional pairwise similarity learning and our central similarity learning. We visualize the high-dimensional Hamming space in the 2d Euclidean space. Each point represents a hash code of an image in the Hamming space, generated by a hash function $f$. Points with the same color denote hash codes of similar pairs. The distance between points reflects the Hamming distance. 
}
\label{fig:motivation}
\vspace{-5pt}
\end{center}
\end{figure}
\fi

\begin{figure}[t!]
\begin{center}
\subfigure[Doublet(pairwise)]{
\includegraphics[scale=0.345]{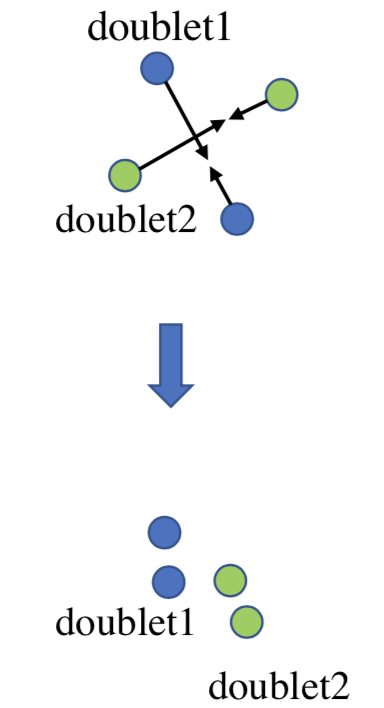}
}
\subfigure[Triplet]{
\includegraphics[scale=0.345]{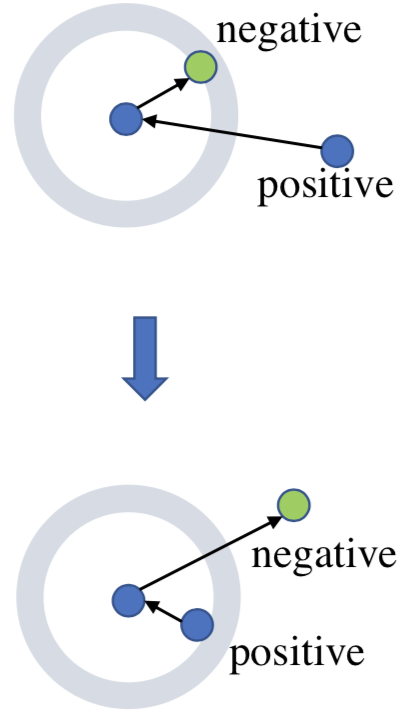}
}
\subfigure[Central similarity]{
\includegraphics[scale=0.35]{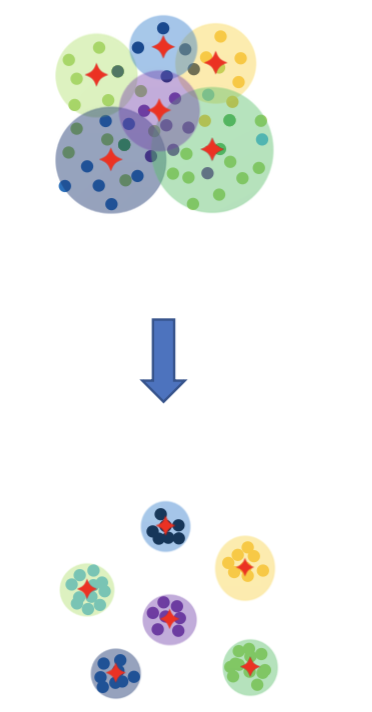}
}
\caption{The intuition behind pairwise/triplet similarity based hashing methods and the proposed center similarity quantization. Pairwise and triplet learnings only consider a pair/triplet of data at once, while our central similarity encourages all similar data points to collapse to the corresponding hash centers (red stars).}
\label{fig:comparison}
\vspace{-5pt}
\end{center}
\vspace{-20pt}
\end{figure}

\begin{figure*}
\begin{center}
\subfigure[Doublet (pairwise similarity)]{
\includegraphics[scale=0.07]{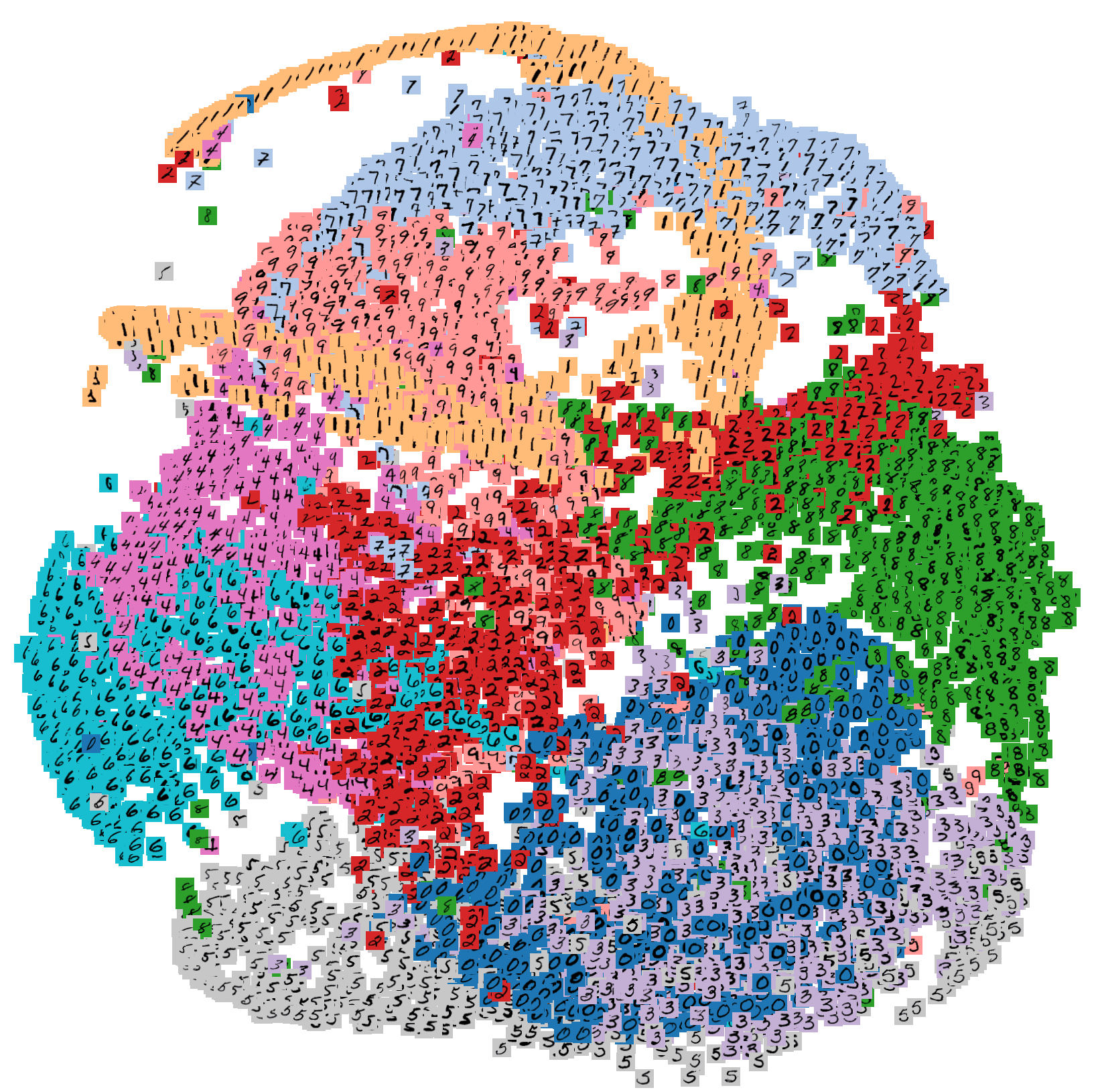}
}
\subfigure[Triplet (triplet similarity)]{
\includegraphics[scale=0.20]{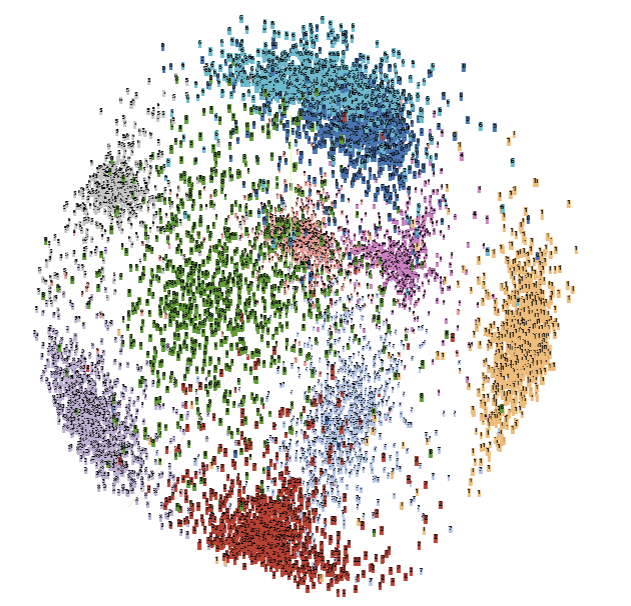}
}
\subfigure[HashNet]{
\includegraphics[width=0.22\textwidth]{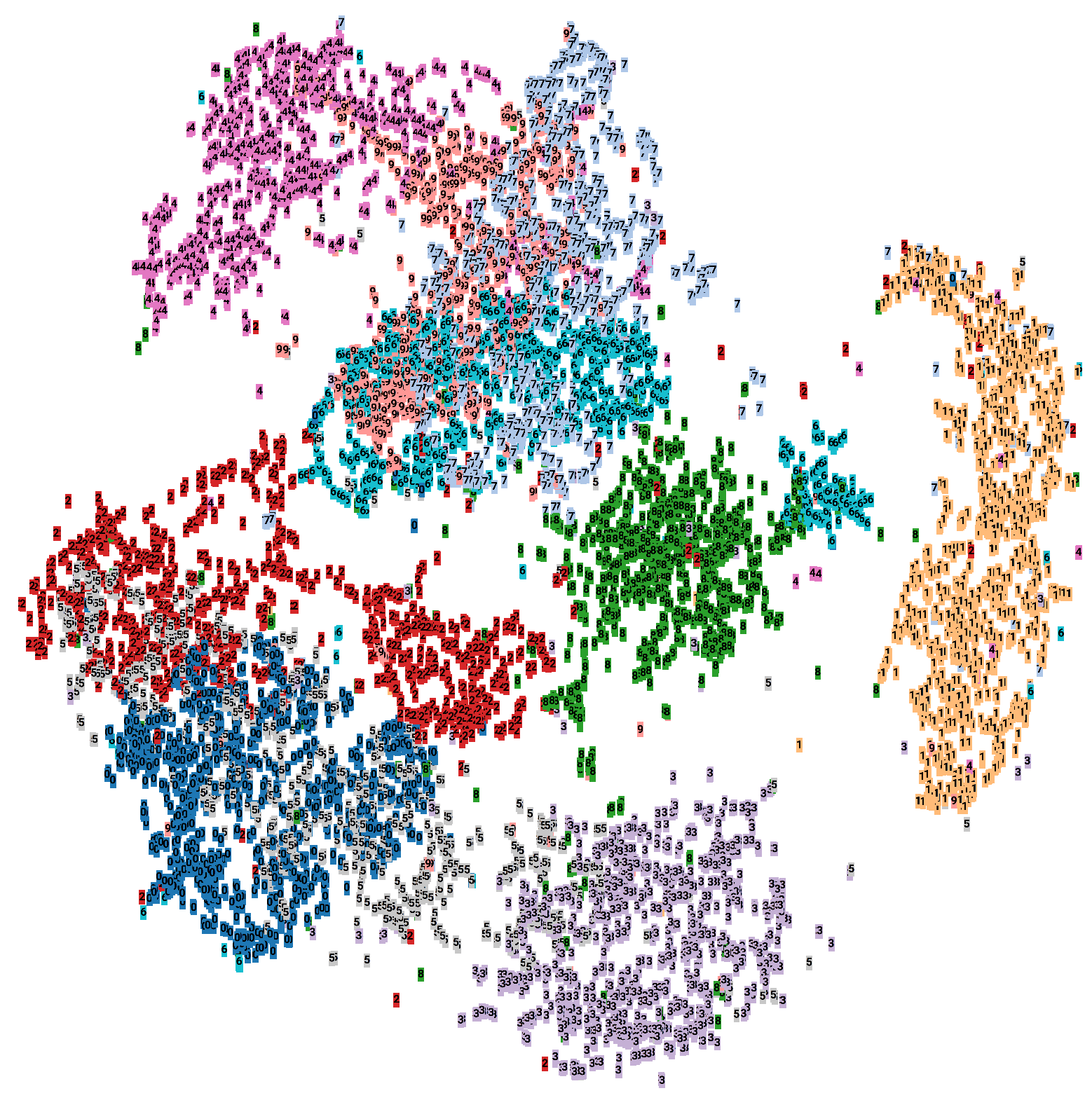}
}
\subfigure[Central similarity (ours)]{
\includegraphics[width=0.23\textwidth]{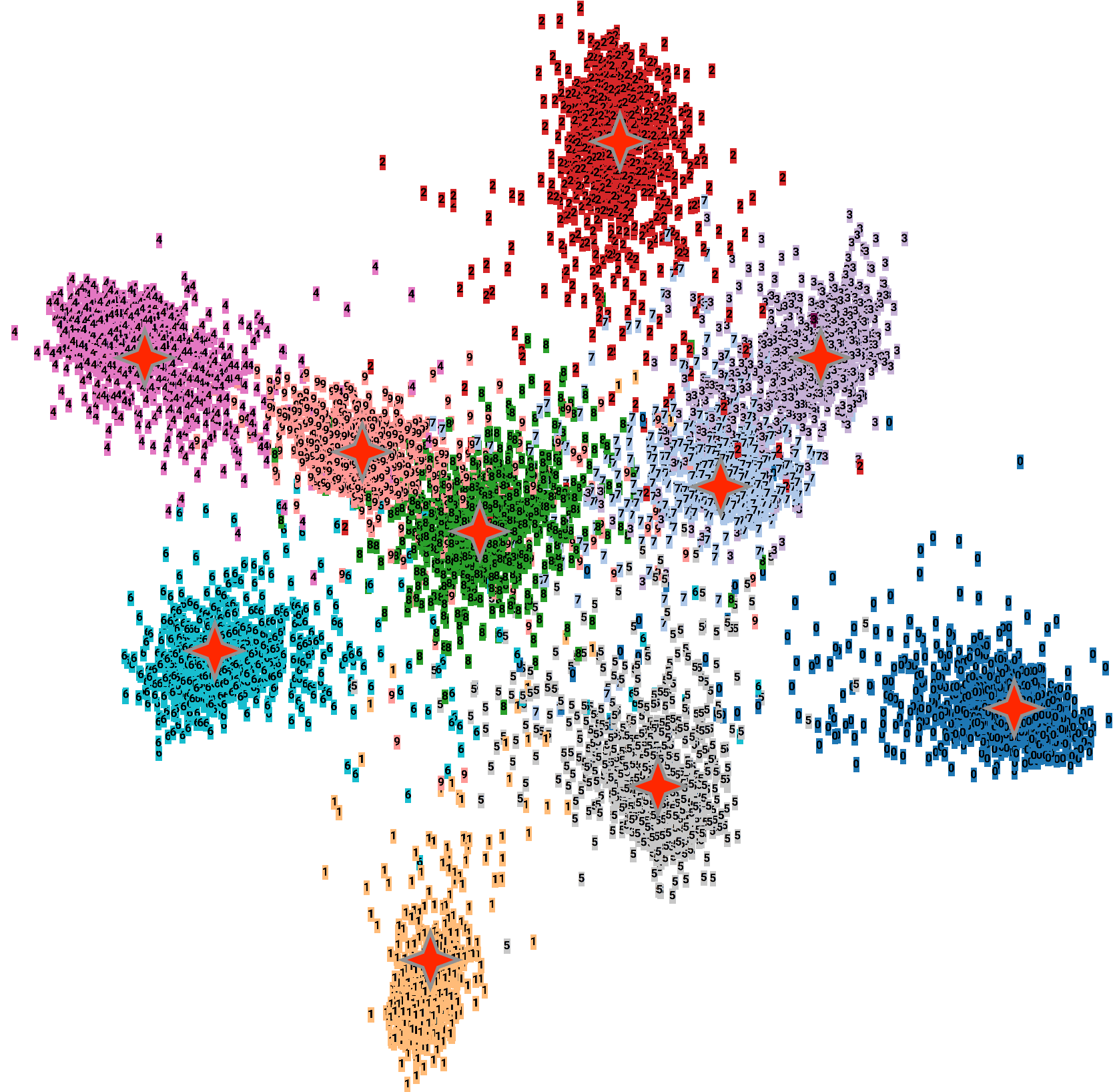}
}
\caption{Visualization of hash codes generated by four deep hash learning methods using different data similarity metrics, trained on MNIST with ten groups of data points (one class for a group). The first two methods use pairwise and triplet similarity respectively. HashNet~\cite{cao2017hashnet} adopts weighted pairwise similarity.}
\label{fig:feature_visualization}
\end{center}
\vspace{-26pt}
\end{figure*}

%\textcolor{red}{[JS; our title is "large-scale" image and video retrieval. But in introduction and contributions, we never mentioned such feature.]}
By transforming high-dimensional data to compact binary hash codes in the Hamming space via a proper hash function~\cite{wang2016learning}, hashing offers remarkable efficiency for data storage and retrieval. 
Recently, ``deep learning to hash" methods~\cite{lai2015simultaneous,shen2015supervised,liong2017deep,li2015feature,liu2016deep,xia2014supervised} have been successfully applied to large-scale image retrieval~\cite{xia2014supervised, zhu2016deep} and video retrieval~\cite{gu2016supervised,qin2017fast,liong2017deep}, which can naturally represent a nonlinear hash function for producing hash codes of input data. 

Most deep hashing methods~\cite{cao2017hashnet, zhu2016deep, norouzi2012hamming, li2015feature} learn hash functions by utilizing pairwise or triplet data similarity, where the data relationships are captured from a local perspective. Such pairwise/triplet based hash learning intrinsically leads to the following issues. 1) Low-efficiency in profiling similarity among the whole training dataset. The commonly used pairwise similarity~\cite{cao2017hashnet,zhu2016deep,li2015feature} or triplet similarity metrics~\cite{norouzi2012hamming,lai2015simultaneous} have a time complexity at an order of $\mathcal{O}(n!)$ for $n$ data points. Thus, it is impractical to exhaustively learn from all the possible data pairs/triplets for large-scale image or video data. 2) Insufficient coverage of data distribution. Pairwise/triplet similarity based methods utilize only partial relationships between data pairs, which may harm the discriminability of the generated hash codes. 3) Low effectiveness on imbalanced data. In real-world scenarios, the number of dissimilar pairs is much larger than that of similar pairs. Hence, pairwise/triplet similarity based hashing methods cannot learn similarity relationships adequately to generate sufficiently good hash codes, leading to restricted performance.

To address the above issues, we propose a new global similarity metric, termed as \emph{central similarity}, which we optimize constantly for obtaining better hash functions. Specifically, the central similarity measures Hamming distances between hash codes and the hash center which is defined as a set of points in the Hamming space with a sufficient mutual distance. Central similarity learning aims at encouraging the generated hash codes to approach the corresponding hash center. With a time complexity of only $\mathcal{O}(nm)$ for $n$ data points and $m$ centers, central similarity based hashing is highly efficient and can generate discriminative enough hash codes from the global data distribution (Fig.~\ref{fig:comparison}), which overcomes the limitations of hashing methods based on the pairwise/triplet similarity. Even in the presence of severe data imbalance, the hash functions can still be well learned from global relationships.

To obtain suitable hash centers, we propose two systematic approaches. One is to directly construct hash centers with maximal mutual Hamming distance by leveraging the Hadamard matrix; the other is to generate hash centers by randomly sampling from Bernoulli distributions. We prove that both approaches can generate proper hash centers that are separated from each other with a sufficient Hamming distance. We also consider jointly learning  hash centers {from data  with hash functions.}
%quanhong: note you mention two methods, but in abstract, only one. And I do not understand red words, pls explain. Remember to mark all changes.
However, we empirically find that learned centers by some common methods~\cite{karlinsky2019repmet,wen2016discriminative,rippel2015metric} cannot provide better hash functions than the analytically constructed ones.
%quanhong: so what are these common methods, belong to which one of your two methods?
We present comparisons on the hash centers from different methods in Sec.~\ref{sec:learn hash center}.

With the generated hash centers, we develop the central similarity with Convolutional Neural Networks (CNNs), to learn a deep hash function in the Hamming space. We name the proposed hashes learning approach as Central Similarity Quantization (CSQ). In particular, we adopt convolutional layers for learning data features and a hash layer for yielding hash codes. After identifying the hash centers, we train the deep CNN and the hash layer end-to-end to generate hash codes with the goal of optimizing central similarity. CSQ is generic and applicable to learning hash codes for both images and videos.

We  conduct illustrative experiments on MNIST~\cite{lecun1998mnist} at first to validate the effectiveness of our CSQ. We find that the hash codes learned by CSQ show favorable intra-class compactness and inter-class separability compared with other state-of-the-art hashing methods, as shown in Fig.~\ref{fig:feature_visualization}. Then we perform extensive comparative experiments on three benchmark datasets for image hashing and two video datasets for video hashing respectively. With CSQ, noticeable improvements in retrieval performance are achieved, \ie, 3\%-20\% in mAP, also with a 3 to 5.5 $\times$ faster training speed over the latest methods.

Our contributions are three-fold. 1) We rethink data similarity modeling and propose a novel concept of hash center for capturing data relationships more effectively. We  present two systematic methods to generate proper hash centers rapidly.  2) We introduce a novel central similarity based  hashing method. It can capture the global data distribution and generate high-quality hash functions efficiently.  To our best knowledge, this is the first work to utilize global  similarity and hash centers for deep hash function learning. 3) We present a deep learning model to implement our method  for  both image and video retrieval and establish new state-of-the-arts.

\section{Related Work}

The ``deep learning to hash" methods such as CNNH~\cite{xia2014supervised}, DNNH~\cite{lai2015simultaneous}, DHN~\cite{zhu2016deep}, DCH~\cite{cao2018deep} and HashNet~\cite{cao2017hashnet} have been successfully applied to image hashing. They adopt 2D CNNs to learn image features and then use hash layers to learn hash codes. Recent hashing methods for images focus on how to design a more efficient pairwise-similarity loss function. DNNH~\cite{lai2015simultaneous} proposes to use a triplet ranking loss for similarity learning. DHN~\cite{zhu2016deep} uses Maximum a Posterior (mAP) estimation to obtain the pairwise similarity loss function. HashNet~\cite{cao2017hashnet} adopts the Weighted Maximum Likelihood (WML) estimation to alleviate the severe the data imbalance by adding weights in pairwise loss functions. Different from previous works~\cite{weiss2009spectral, liu2011hashing, liu2014discrete,liu2016multimedia, wang2013learning, shen2015learning, shen2015supervised,li2017sub,song2015top, li2018self,li2019cross,yang2019distillhash}, this work proposes a new central similarity metric and use it to model the relationships between similar and dissimilar pairs for improving the discriminability of generated hash codes.

Compared with image analysis, video analysis aims to utilize the temporal information~\cite{simonyan2014two, donahue2015long, wang2016temporal, varol2017long, chao2018rethinking, yuan2019cycle, yuan2019unsupervised}. Video hashing methods such as DH~\cite{qin2017fast}, SRH~\cite{gu2016supervised}, DVH~\cite{liong2017deep} exploit the temporal information in videos compared with image hashing. For instance,~\cite{qin2017fast} utilizes Disaggregation Hashing to exploit the correlations among different feature dimensions.~\cite{gu2016supervised} presents an LSTM-based method to capture the temporal information between video frames. Recently,~\cite{liong2017deep} fuses the temporal information by using fully-connected layers and frame pooling. Different from these hashing methods, our proposed CSQ is a generic method for both image and video hashing. Through directly replacing 2D CNNs with 3D CNNs, the proposed CSQ can well capture the temporal information for video hashing.

Our CSQ is partially related to center loss in face recognition~\cite{wen2016discriminative} which uses a center loss to learn more discriminative representation for face recognition (classification). The centers in~\cite{wen2016discriminative} are derived from the feature representation of the corresponding categories, which are unstable with intra-class variations. Different from this center loss for recognition~\cite{wen2016discriminative}, our proposed hash center is defined over hash codes instead of feature representations, and can help generate high-quality hash codes in the Hamming space.

\section{Method}
We consider learning a hash function in a supervised manner from a training set of $N$ data points $\mathcal{X} = \left \{ \{x_{i}\}_{i=1}^{N}\,,\, L \right \}$, where each $x_i\in\mathbb{R}^{D}$ is a data point to hash and $L$ denotes the semantic label set for data $\mathcal{X}$. Let $f:x\mapsto h\in \left \{ 0,1 \right \}^{K}$ denote the nonlinear hash function from the input space $\mathbb{R}^{D}$ to $K$-bit Hamming space $\left \{ 0,1 \right \}^{K}$. Similar to other supervised ``deep learning to hash" methods~\cite{cao2017hashnet,zhu2016deep}, we pursue a hash function that is able to generate hash codes $h$'s for the data points $x$'s which are close in the Hamming space and share similar semantic labels.

We define a set of points $\mathcal{C}=\{c_1, c_2,\ldots, c_m\} \subset \{0,1\}^K$  with a sufficient distance in the Hamming space as \emph{hash centers},
%quanhong: i still do not know whether a center corresponds to a point, or a set of points. Yuan: a hash center is a point, hash centers are a set of point.
and propose to learn hash functions supervised by the central similarity w.r.t.\ $\mathcal{C}$. The central similarity would encourage similar data pairs to be close to a common hash center and dissimilar data pairs to be distributed around different hash centers respectively. Through such central similarity learning, the global similarity information between data pairs can be preserved in $f$, yielding high-quality hash codes.

In below, we first give a formal definition of hash center and explain how to generate proper hash centers systematically. Then we elaborate on the details of the central similarity quantization. 

\vspace{-5pt}

\subsection{Definition of Hash Center}
The most intuitive motivation is to learn hash centers from image or video features, such that the learned centers preserve ``distinctness'' between different data points. However, we find that hash centers learned from data features with diverse mutual Hamming distance do not perform better than hash centers with pre-defined Hamming distance (in Experiments, Sec~\ref{sec:learn hash center}). We thus assume that each center should be more distant from the other centers than to the hash codes associated with it. As such, the dissimilar pairs can be better separated and similar pairs can be aggregated cohesively. Based on the observation and intuition, we formally define a set of points in the Hamming space as valid \emph{hash centers} with the following properties.
\begin{definition}[Hash Center]
\label{def:hash_center}
We define hash centers as a set of points $\mathcal{C}=\left \{ c_{i}\right \}_{i=1}^m \subset \left \{ 0, 1 \right \}^{K} $ in the $K$-dimensional Hamming space with an average pairwise distance satisfying% $\frac{K}{2}$: 
\begin{equation}
\begin{aligned}
    \frac{1}{T}\sum_{ i\neq j}^m D_{H}\left ( c_{i},c_{j} \right ) \geqslant \frac{K}{2},
\end{aligned}
\label{eqn:hash_center}
\end{equation}
where $D_{H}$ is the Hamming distance, $m$ is the number of hash centers, and $T$ is the number of combinations of different $c_i$ and $c_j\in\mathcal{C}$.
\end{definition}
%quanhong: still, does this formulation tell that a hash center is a point?

\begin{figure}[h!]
\begin{center}
\subfigure[3d Hamming space]{
\label{subfig:3d_ham}
\includegraphics[width=0.3\linewidth]{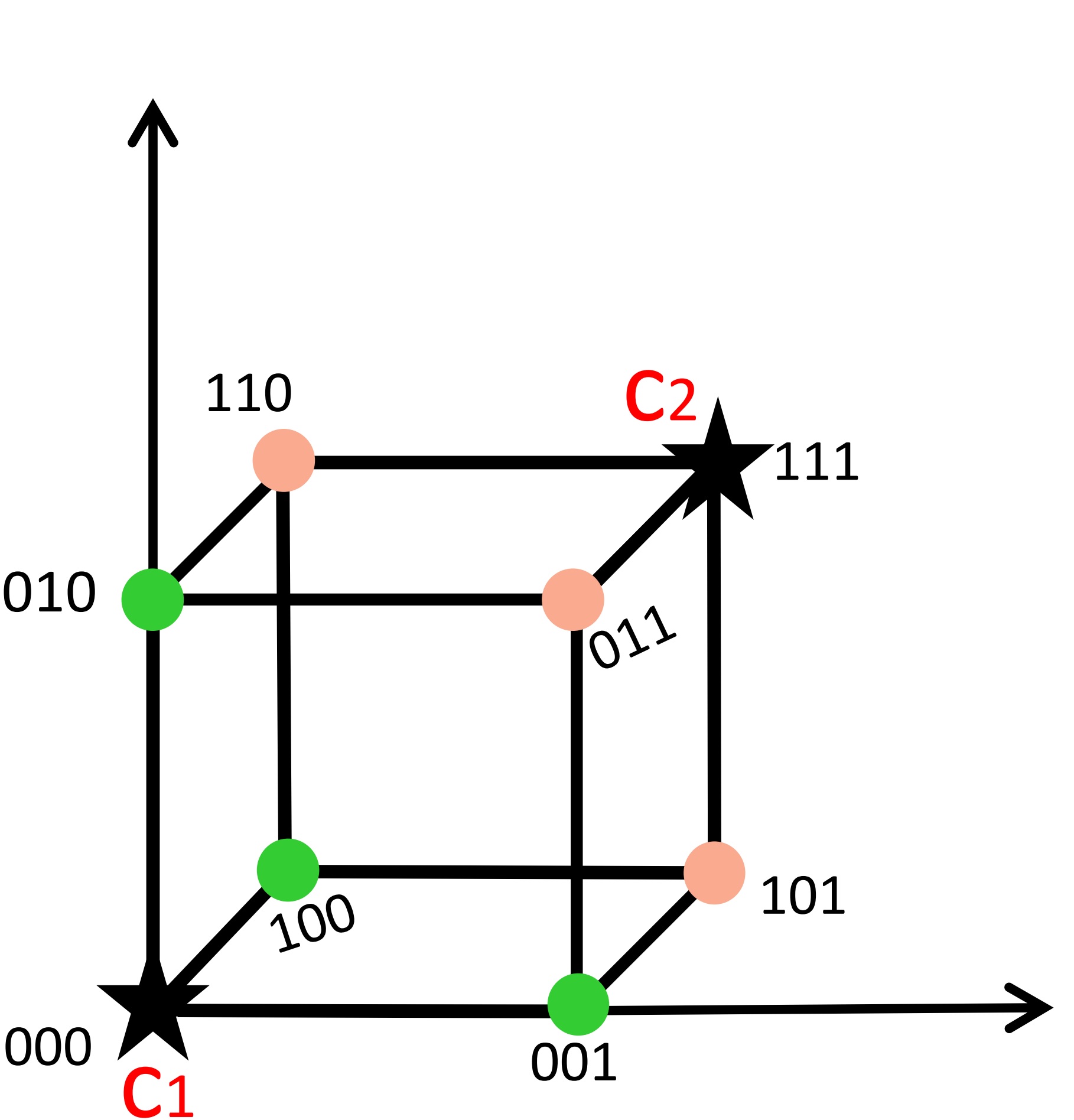}
}
\subfigure[4d Hamming space]{
\label{subfig:4d_ham}
\includegraphics[width=0.5\linewidth]{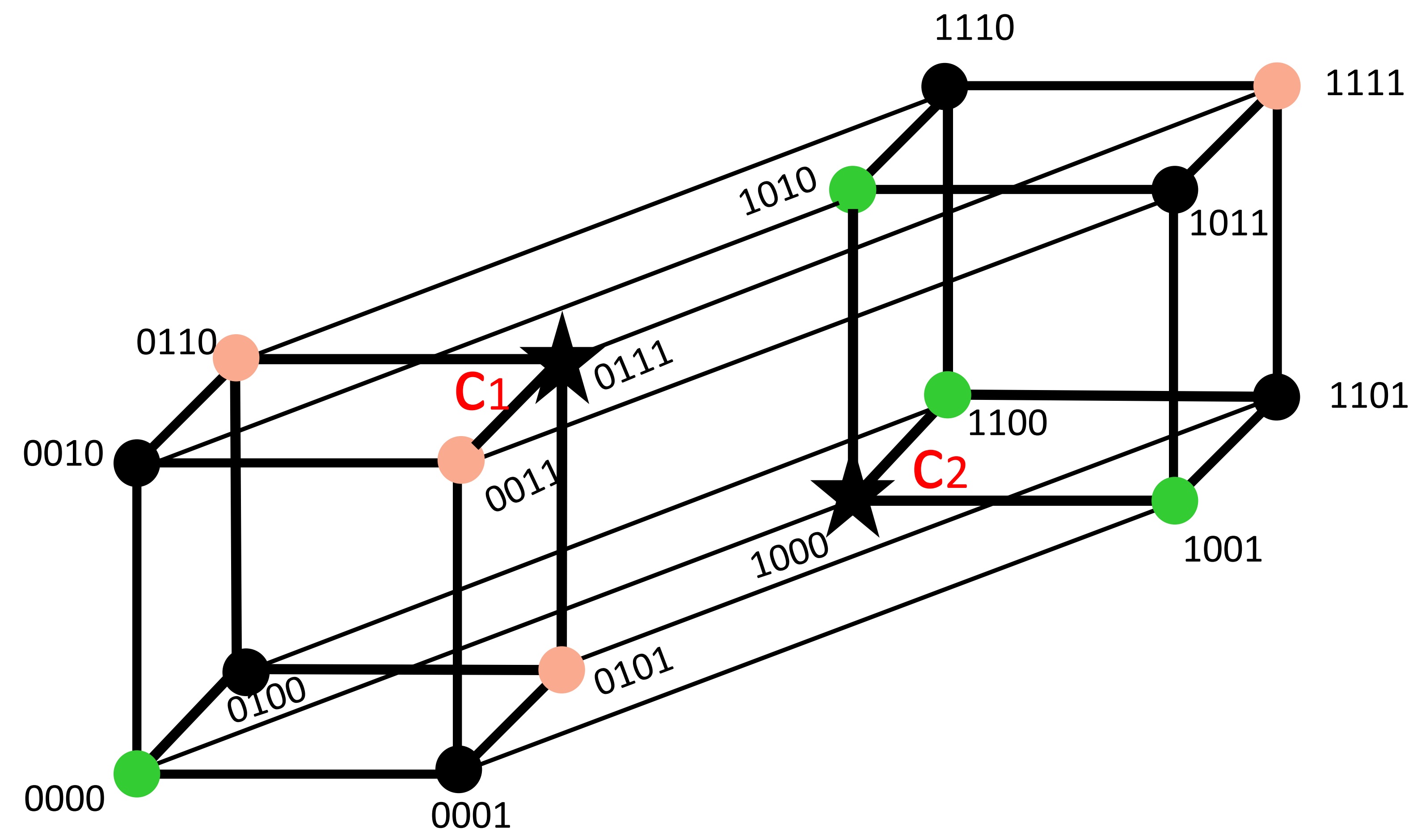}
}
\caption{The illustration of hash centers in 3d and 4d Hamming spaces.}
\label{fig:hash-center}
\end{center}
\vspace{-20pt}
\end{figure}

For better clarity, we show some examples
of the desired hash centers in the 3d and 4d Hamming spaces in Fig.~\ref{fig:hash-center}. In Fig.~\ref{subfig:3d_ham}, the hash center of hash codes $[0,1,0]$, $[0,0,1]$ and $[1,0,0]$ is $c_1$, and the Hamming distance between $c_1$ and $c_2$ is 3.
%quanhong: one question that may not be important: why '3' with no unit?
In Fig.~\ref{subfig:4d_ham}, we use a 4d hypercube to represent the 4d Hamming spaces. The two stars  $c_{1}$ and $c_{2}$ are the hash centers given in Definition~\ref{def:hash_center}. The distance between $c_{1}$ and $c_{2}$ is $D_{H}\left ( c_{1},c_{2} \right ) = 4$, and the distance between the green dots and the center $c_{2}$ is the same ($D_{H}=1$). However, we do not strictly require all points to have the same distance from the corresponding center. Instead, we define the nearest center as the corresponding hash center for a hash code.  

\subsection{Generation of Hash Centers}
We develop two approaches for generating valid hash centers based on the following observation. In the $K$-dimensional Hamming space (throughout this paper, $K$ is set to an even number), if a set of points are mutually orthogonal, they will have an equal distance of $K/2$ to each other. Namely, they are valid hash centers satisfying Definition~\ref{def:hash_center}.

Our first approach is to generate hash centers by leveraging the following nice properties of a Hadamard matrix.  It is known that a $K \times K$ Hadamard matrix $H_{K}=[h_a^1; \ldots ; h_a^K]$ satisfies:
1) It is a squared matrix with rows $h_{a}^{i}$ being mutually orthogonal, \emph{i.e.}, the inner products of any two row vectors $ \langle h_{a}^{i}, h_{a}^{j}  \rangle =0$. The Hamming distance between any two row vectors is $D_{H}(h_{a}^{i},h_{a}^{j})=\frac{1}{2}(K- \langle h_{a}^{i}, h_{a}^{j}  \rangle) = {K}/{2}$. Therefore, we can choose hash centers from these row vectors. 2) Its size $K$ is a power of $2$ (\emph{i.e.},  $K=2^{n}$), which is consistent with the customary number of bits of hash codes. 3) It is a binary matrix whose entries are either -1 or +1. We can simply replace all -1 with 0 to obtain  hash centers in $\left \{ 0,1 \right \}^{K}$. 
%need revise this paragraph

To sample the hash centers from the Hadamard matrix, we first build a $K \times K$ Hadamard matrix by  Sylvester's construction~\cite{weisstein2002hadamard} as follows:
\begin{equation}
\centering
H_{K}=\begin{bmatrix}
H_{2^{n-1}} &H_{2^{n-1}} \\ 
H_{2^{n-1}} &-H_{2^{n-1}} 
\end{bmatrix}=H_{2}\otimes H_{2^{n-1}},
\end{equation}
where $\otimes$ represents the Hadamard product, and $K=2^{n}$. The two factors within the initial Hadamard matrix are $H_{1}=\begin{bmatrix} 1 \end{bmatrix}$ and $H_{2}=\begin{bmatrix}1 &1 \\ 1 &-1 \end{bmatrix}$. When the number of centers $m\leqslant K$, we directly choose each row to be a hash center. 
When $K<m\leqslant 2K$, we use a combination of two Hadamard matrices $H_{2K}=[H_{K}, -H_{K}]^\top$ to construct hash centers. %\footnote{Proof that $H_{2K}$ can be valid hash centers in supplementary material.}.

\begin{algorithm}[t!]\footnotesize
		\caption{Generation of Hash Centers}
		\SetKwInOut{Input}{Input} \SetKwInOut{Output}{Output}
		\Input{The number of hash centers $m$, the dimension of the Hamming space (hash codes) $K$.}
		\textbf{Initialization:}  construct a $K\times K$ Hadamard matrix $H_{K}=\left [ h_{a}^{i} \right ]$ and construct $H_{2K}=[H_{K}, -H_{K}]^\top = \left [h_{2k}^{i} \right ].$ \\
		\For{iteration $i$, $i{=}1$ to $m$}{
			\If(\qquad\tcp*[h]\,{$n$ is any $\mathbb{Z}^{+}$}){$m\leqslant K$ \& $K=2^{n}$}{$c_{i}=h_{a}^{i}$;} 
			\uElseIf{$K<m\leqslant2K$ \& $K=2^{n}$ }{$c_{i}=h_{2k}^{i}$;}
			\Else{ $c_{i}[\text{random half position}]=1$;\\
			$c_{i}[\text{other half position}]=0$;}
		}
		Replace all -1 with 0 in these centers;\\
		\Output{hash centers: $\mathcal{C}=\left \{ c_1,\ldots,c_m \right \} \subset \{0,1\}^K$.}
\label{alg1}
\vspace{-2pt}
\end{algorithm}

Though applicable in most cases, the number of valid centers generated by the above approach is constrained  by the fact that the Hadamard matrix is a squared one. If $m$ is larger than $2K$ or $K$ is not the power of 2, the first approach is inapplicable. We thus propose the second generation approach by randomly sampling the bits of each center vector. In particular, each bit of a center $c_i$ is sampled from a Bernoulli distribution $\mathrm{Bern(0.5)}$ where  $P(x=0)=0.5$ if $x\sim \mathrm{Bern(0.5)} $. We can easily prove that the distance between these centers is $K/2$ in expectation. Namely, $\mathbb{E}[D_H(c_i,c_j)] = K/2$ if $c_i,c_j \sim \mathrm{Bern(0.5)}$. We summarize these two approaches in Alg.~\ref{alg1}. The generation algorithm is very efficient and only needs a trivial computation/time cost to generate hash centers.

Once a set of hash centers is obtained, the next step is to associate the training  data samples $\mathcal{X}$ with their individual corresponding centers to compute the central similarity. Recall $L$ is the semantic label for $\mathcal{X}$, and usually $L=\{l_1,\ldots, l_q\}$. For single-label data, each data sample belongs to one category, while each multi-label data sample belongs to more than one category. We term the hash centers that are generated from Alg.~\ref{alg1} and associated with semantic labels as \emph{semantic hash centers}. We now explain how to obtain the semantic hash centers for single- and multi-label data separately. 
\vspace{-10pt}

\paragraph{Semantic hash centers for single-label data}
For single-label data, we assign one hash center for each category. That is, we generate $q$ hash centers $\{c_1,\ldots, c_{q}\}$ by Alg.~\ref{alg1} corresponding to labels $\{l_1,\ldots, l_{q}\}$. Thus, data pairs with the same label share a common center and are encouraged to be close to each other. Because each data sample is assigned to one hash center, we obtain the semantic hash centers $\mathcal{C'} = \{c'_1, c'_2,\ldots, c'_N\}$, where $c'_i$ is the hash center of $x_i$.
\label{subsec:single-label}
\vspace{-20pt}

\paragraph{Semantic hash centers for multi-label data}

\begin{figure}[h!]
\begin{center}
\setlength{\tabcolsep}{1.5pt}
\includegraphics[scale=0.03]{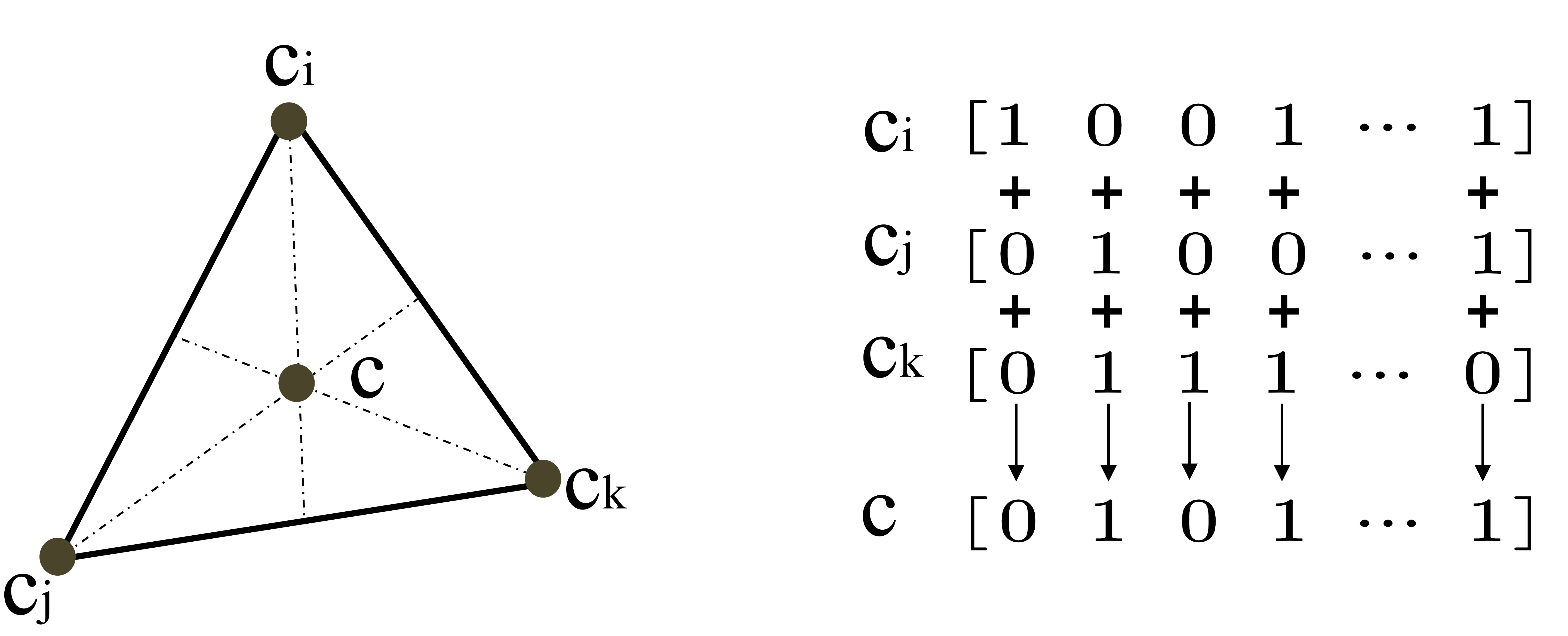}
\caption{Semantic hash center for multi-label data.}
\label{fig:center_multi_label}
\end{center}
\vspace{-15pt}
\end{figure}

For multi-label data, DCH~\cite{cao2018deep}, HashNet~\cite{cao2017hashnet} and DHN~\cite{zhu2016deep} directly make data pairs similar if they share at least one category. However, they ignore the transitive similarity when data pairs share more than one category. In this paper, we generate transitive centers for data pairs sharing multiple labels. First, we generate $q$ hash centers $\{c_1,\ldots, c_{q}\}$ by Alg.~\ref{alg1} corresponding to semantic labels $\{l_1,\ldots, l_{q}\}$. Then for data including two or more categories, we calculate the centroid of these centers, each of which corresponds to a single category.  
For example, suppose one data sample $x\in \mathcal{X}$ has three categories $l_{i}$, $l_{j}$ and $l_{k}$. The centers of the three categories are $c_{i}$, $c_{j}$ and $c_{k}$, as shown in Fig.~\ref{fig:center_multi_label}. We calculate the centroid $c$ of the three centers as the hash center of $x$. To ensure the elements to be binary, we calculate each bit by %\textcolor{red}{
voting at the same bit of the three centers and taking the value that dominates, as shown in the right panel of Fig.~\ref{fig:center_multi_label}. %\textcolor{red}{
If the number of 0 is equal to the number of 1 at some bits (\emph{i.e.}, the voting result is a draw), we sample from $\mathrm{Bern(0.5)}$ for these bits. Finally, for each $x_i\in\mathcal{X}$, we take the centroid as its semantic hash center, and then obtain semantic hash centers $\mathcal{C'}=\{c'_1, c'_2,\ldots, c'_N\}$, where $c'_i$ is the hash center of $x_i$.
\label{subsec:multi-label}

\subsection{Central Similarity Quantization}
\label{central similarity quantization}
Given the generated centers $\mathcal{C} = \{c_1,\ldots,c_q\}$ for training data $\mathcal{X}$ with $q$ categories, we obtain the semantic hash centers $\mathcal{C'}=\{c'_1, c'_2,\ldots, c'_N\}$ for single- or multi-label data, where $c'_i$ denotes the hash center of the data sample $x_i$. We derive the central similarity learning objective by maximizing the logarithm posterior of the hash codes w.r.t.\ the semantic hash centers. Formally, the logarithm Maximum a Posterior (MAP) estimation of hash codes $\mathcal{H}=\left [ h_{1},...,h_{N} \right ]$ for all the training data can be obtained by maximizing the following likelihood probability: 
\vspace{-5pt}
\begin{equation*}
\log P(\mathcal{H}|\mathcal{C'})\propto\log P(\mathcal{C'}|\mathcal{H})P(\mathcal{H})=\sum_{i}^{N}\log P(c'_{i}|h_{i})P(h_{i}),
\vspace{-10pt}
\end{equation*}
where $P(\mathcal{H})$ is the prior distribution over hash codes and $P(\mathcal{C'}|\mathcal{H})$ is the likelihood function. $P(c'_{i}|h_{i})$ is the conditional probability of center $c'_i$ given hash code $h_i$.  We model $P(\mathcal{C'}|\mathcal{H})$ as a Gibbs distribution: $P(c'_i|h_i)=\frac{1}{\alpha }\exp(-\beta D_H(c'_i,h_i))$, where $\alpha$ and $\beta$ are constants, and $D_H$ measures the Hamming distance between a hash code and its hash center. Since hash centers are binary vectors, we use Binary Cross Entropy (BCE) to measure the Hamming distance between the hash code and its center, $D_H(c'_i,h_i)=\text{BCE}(c'_i,h_i)$. So the conditional probability is calculated as $\log P(c'_{i}|h_{i}) \propto -\frac{1}{K}\sum_{k\in K} ( c'_{i,k}\log h_{i,k}+(1-c'_{i,k})\log(1-h_{i,k}) )$. We can see that the larger the conditional probability $P(c'_i|h_i)$ is, the smaller the Hamming distance will be between hash code $h$ and its hash center $c$,  implying that the hash code is close to its corresponding center; otherwise the hash code is far away from its corresponding center.
By substituting $\log P(c'_{i}|h_{i})$ into the MAP estimation, we obtain the optimization objective of the central similarity loss $L_{C}$:
\begin{equation}
\label{eqn:central_similarity}
\begin{aligned}
    L_{C}=&\frac{1}{K}\sum_{i}^{N}\sum_{k\in K} \left [c'_{i,k}\log h_{i,k}+(1-c'_{i,k})\log(1-h_{i,k}) \right ].\\
\end{aligned}
\end{equation}

Since each hash center is binary, existing optimization cannot guarantee that the generated hash codes completely converge to hash centers~\cite{weise2009optimization} due to the inherent optimization difficulty. So we introduce a quantization loss $L_{Q}$ to refine the generated hash codes $h_{i}$. Similar to DHN~\cite{zhu2016deep}, we use the bi-modal Laplacian prior for quantization, which is defined as $L_{Q}=\sum_{i\neq j}^{N}(|| |2h_{i}-\textbf{1}|-\textbf{1}||_{1})$,
where $\textbf{1}\in\mathbb{R}^{K}$ is an all-one vector.
As $L_{Q}$ is a non-smooth function which makes it difficult to calculate its derivative, we adopt the smooth function $\log \cosh$~\cite{aaponatural} to replace it. So $|x|\approx \log \cosh x$. Then the quantization loss $L_{Q}$ becomes
\begin{equation}
\begin{aligned}
    L_{Q}=&\sum_{i}^{N}\sum_{k=1}^{K}(\log \cosh(|2h_{i,k}-1|-1).
\end{aligned}
\vspace{-5pt}
\end{equation}
Finally, we have central similarity optimization problem:
%quanhong: why say problem? I think 'objective' is better.
\begin{equation}
    \min_{\Theta } L_{T}=L_{C} + \lambda_1 L_{Q}
\label{eqn: total_objective}
\vspace{-5pt}
\end{equation}
where $\Theta$ is the set of all parameters for deep hash function learning, and $\lambda_1$ is the hyper-parameter obtained through grid search in our work\footnote{We provide formulation for jointly estimating central similarity and pairwise similarity to learn deep hash functions in supplementary material.}.

Based on the loss function $L_{T}$, we adopt the standard framework~\cite{he2016deep, cao2017hashnet} in deep-hashing methods to conduct CSQ. Specifically, multiple convolutional layers are adopted to learn data features and a hash layer with three $fc$ layers and ReLU as the activation function is used to generate hash codes. The detailed framework of CSQ is given in the supplementary material.

\iffalse
\subsection{Architecture of CSQ}
Base on these definitions and designs, we propose a Hash Center Network (CSQ) to learn central similarity for image and video hashing. The network architecture is shown in Fig.~\ref{fig:framework}. The input of CSQ is $\left \{ (x_{i}, x_{j}, c_{i}, c_{j}) \right \}$.  
Here $c_{i}$ and $c_{j}$ are the hash centers for $x_{i}$ and $x_{j}$ respectively. CSQ takes this input and outputs compact hash codes through the following deep hashing pipeline: 1) a 2D or 3D CNN sub-network to extract the data representation for image or video data, 2) a hash layer with three fully-connected layers and activation functions to project high dimensional data features to hash codes in the Hamming space, 3) a central similarity loss $L_{C}$ for central similarity-preserving learning, where all hash centers are defined in the Hamming space, making hash codes converge on corresponding centers.  and 4) a quantization loss $L_{Q}$ for improving binarization. 
\fi

\section{Experiments}
We conduct experiments for both image and video retrieval to evaluate our CSQ against several state-of-the-arts. Five benchmark (image and video) datasets are used in our experiments and their statistics are summarized in Tab.~\ref{tab:datasets}.

\begin{table}[h]
\small
\begin{center}
\fontsize{8pt}{12pt}\selectfont
\caption{Experimental settings for all datasets. DI (Data Imbalance) is ratio between the number of dissimilar and similar pairs.}
\begin{tabular}{l|c|c|c|c|c}
%{|p{1.3cm}|c|c|p{0.8cm}|p{1cm}|p{0.5cm}|}
\toprule
Dataset&Data Type&\#Train&\#Test&\#Retrieval&DI\\
\midrule ImageNet&image&10,000&5,000&128,495& 100:1\\
 MS COCO&image&10,000&5,000&112,217&1:1\\
 NUS\_WIDE&image&10,000&2,040&149,685&5:1\\
 UCF101 &video &9.5k &3.8k &9.5k &101:1\\
 HMDB51&video &3.5k &1.5k &3.5k&51:1\\
\bottomrule
\end{tabular}
\label{tab:datasets}
\end{center}
\vspace{-25pt}
\end{table}

\begin{table*}
\small
\begin{center}
\fontsize{8pt}{11pt}\selectfont
\setlength\tabcolsep{9pt}
%\captionsetup{justification=centering}
\caption{Comparison in mAP of Hamming Ranking for different bits on image retrieval.}
\begin{tabular}{l|c|c|c|c|c|c|c|c|c}
\toprule
\multirow{2}{*}{Method} & \multicolumn{3}{c|}{ImageNet (mAP@1000)} & \multicolumn{3}{c|}{MS COCO (mAP@5000)} & \multicolumn{3}{c}{NUS-WIDE (mAP@5000)} \\
\cline{2-10}
~ & 16 bits & 32 bits & 64 bits & 16 bits & 32 bits & 64 bits & 16 bits & 32 bits & 64 bits \\

\midrule
ITQ-CCA~\cite{gong2013iterative} &0.266 &0.436 &0.576 &0.566 &0.562&0.502 &0.435&0.435&0.435\\
 BRE~\cite{kulis2009learning} &0.063&0.253&0.358 &0.592&0.622&0.634 &0.485&0.525&0.544 \\
 KSH~\cite{liu2012supervised}&0.160&0.298&0.394  &0.521&0.534&0.536 &0.394&0.407&0.399\\
 SDH~\cite{shen2015supervised} &0.299&0.455&0.585 &0.554&0.564&0.580 &0.575&0.590&0.613 \\
 
 \midrule
 CNNH~\cite{xia2014supervised} &0.315&0.473&0.596 &0.599&0.617&0.620 &0.655&0.659&0.647 \\
 DNNH~\cite{lai2015simultaneous} &0.353&0.522&0.610 &0.644&0.651&0.647 &0.703&0.738&0.754 \\
 DHN~\cite{zhu2016deep} &0.367&0.522&0.627  &0.719&0.731&0.745    &0.712&0.759&0.771 \\
 HashNet~\cite{cao2017hashnet} &0.622&0.701&0.739 &0.745&0.773&0.788 &0.757&0.775&0.790 \\
 DCH~\cite{cao2018deep} &0.652&0.737&0.758&0.759&0.801&0.825&0.773&0.795&0.818\\
 
\midrule \textbf{CSQ (Ours)} &\textbf{0.851}&\textbf{0.865}&\textbf{0.873} &\textbf{0.796}&\textbf{0.838}&\textbf{0.861} &\textbf{0.810}&\textbf{0.825}&\textbf{0.839} \\
\bottomrule
\end{tabular}
\label{tab:MAP_image}
\vspace{-20pt}
\end{center}
\end{table*}

\begin{table}[h]
\small
\begin{center}
\fontsize{8pt}{12pt}\selectfont
\setlength\tabcolsep{5pt}
\caption{Comparison in mAP of Hamming Ranking by adopting different backbones (AlexNet or ResNet50) to learn features.}
\begin{tabular}{l|c|c|c|c|c|c}
\toprule
\multirow{2}{*}{Method} & \multicolumn{3}{c|}{ImageNet (AlexNet)} & \multicolumn{3}{c}{ImageNet (ResNet50)} \\
\cline{2-7}
~ & 16 bits & 32 bits & 64 bits & 16 bits & 32 bits & 64 bits\\
\midrule
{CNNH}~\cite{xia2014supervised} &0.282&0.453&0.548 &0.315&0.473&0.596 \\

{DNNH}~\cite{lai2015simultaneous} &0.303&0.457&0.572 &0.353&0.522&0.610 \\

{DHN}~\cite{zhu2016deep} &0.318&0.473&0.569 &0.367&0.522&0.627 \\
{HashNet}~\cite{cao2017hashnet} &0.506&0.631&0.684 &0.622&0.701&0.739 \\
{DCH}~\cite{cao2018deep} &0.529&0.637&0.664&0.652&0.737&0.758\\
\midrule
\textbf{CSQ (Ours)} &\textbf{0.601}&\textbf{0.653}&\textbf{0.695} &\textbf{0.851}&\textbf{0.865}&\textbf{0.873} \\
\bottomrule
\end{tabular}
\label{tab:mAP_2}
\vspace{-15pt}
\end{center}
\end{table}

\begin{table}[h]
\small
\begin{center}
\fontsize{8pt}{12pt}\selectfont
\setlength\tabcolsep{5pt}
\caption{Training time (in mins) comparison on three datasets with different hash bits.(One GPU: TITAN X; Backbone: AlexNet).}
\begin{tabular}{l|c|c|c|c|c|c}
\toprule
\multirow{2}{*}{Method} & \multicolumn{2}{c|}{ImageNet} & \multicolumn{2}{c|}{COCO}& \multicolumn{2}{c}{NUS\_WIDE} \\
\cline{2-7}
~ & 32 bits & 64 bits & 32 bits & 64 bits & 32 bits & 64 bits\\
\midrule
{DHN}~\cite{zhu2016deep} &3.87e2&4.13e2&3.92e2 &4.05e2&3.56e2&3.63e2 \\
%{DCH}~\cite{cao2018deep} &5.44e2&5.37e2&5.29e2 &5.52e2&5.38e2&5.17e2 \\
{HashNet}~\cite{cao2017hashnet} &6.51e2&6.84e2&6.42e2 &6.88e2&7.29e2&7.34e2 \\
\midrule
\textbf{CSQ (Ours)} &\textbf{0.92e2}&\textbf{1.01e2}&\textbf{1.13e2} &\textbf{1.15e2}&\textbf{1.30e2}&\textbf{1.39e2} \\
\bottomrule
\end{tabular}
\label{tab:training_time}
\vspace{-25pt}
\end{center}
\end{table}

\subsection{Experiments on Image Hashing}

\paragraph{Datasets} We use three image benchmark datasets, including ImageNet~\cite{russakovsky2015imagenet}, NUS\_WIDE~\cite{chua2009nus} and MS COCO~\cite{lin2014microsoft}.
On ImageNet, we use the same data and settings as~\cite{cao2017hashnet, zhu2016deep}.  As ImageNet is a single-label dataset, we directly generate one hash center for each category. MS COCO is a multi-label image dataset with 80 categories.
NUS\_WIDE is also a multi-label image dataset, and we choose images from the 21 most frequent categories for evaluation~\cite{zhu2016deep,lai2015simultaneous}. For MS COCO and NUS\_WIDE datasets, we first generate 80 and 21 hash centers for all categories respectively, and then calculate the centroid of the multi-centers as the semantic hash centers for each image with multiple labels, following the approach in Sec.~\ref{subsec:multi-label}. The visualization of generated hash centers is given in the supplementary material. 
\vspace{-5pt}

\paragraph{Baselines and evaluation metrics} We compare retrieval performance of our proposed CSQ with nine classical or state-of-the-art hashing/quantization methods, including four supervised shallow methods
ITQ-CCA~\cite{gong2013iterative}, BRE~\cite{kulis2009learning}, KSH~\cite{liu2012supervised}, SDH~\cite{shen2015supervised} and five supervised deep methods CNNH~\cite{xia2014supervised}, DNNH~\cite{lai2015simultaneous}, DHN~\cite{zhu2016deep}, HashNet~\cite{cao2017hashnet} and DCH~\cite{cao2018deep}. For the four shallow hashing methods, we adopt the results from the latest works~\cite{zhu2016deep, cao2017hashnet, cao2018deep} to make them directly comparable. We evaluate image retrieval performance based on four standard evaluation metrics: Mean Average Precision (mAP), Precision-Recall curves (PR), and Precision curves w.r.t.\ different numbers of returned samples (P@N), Precision curves within Hamming distance 2 (P@H=2). We adopt mAP@1000 for ImageNet as each category has 1,300 images, and adopt mAP@5000 for MS COCO and NUS\_WIDE.

%For fair comparisons, we reproduce or use the official implementations for the five deep hashing methods: CNNH, DNNH, DHN, HashNet and DCH, and adopt the same backbone architecture with our CSQ . All experiments are conducted in the same setting. Hyper-parameters for all methods are obtained by grid search.

%\vspace{-5pt}

\begin{figure*}[h!]
\begin{center}
\subfigure[P-R curve @64bits]{
\includegraphics[scale=0.32]{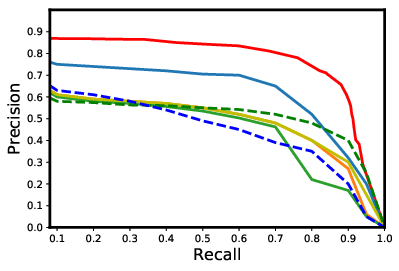}
\label{subfig:image_pr}
}
\subfigure[P@N @64bits]{
\includegraphics[scale=0.32]{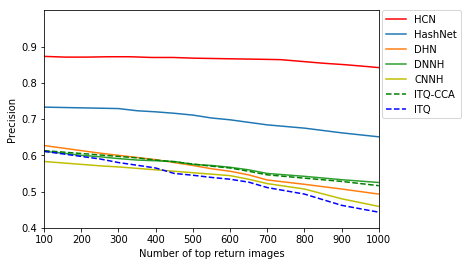}
\label{subfig:image_pn}
}
\subfigure[P@H=2]{
\includegraphics[scale=0.32]{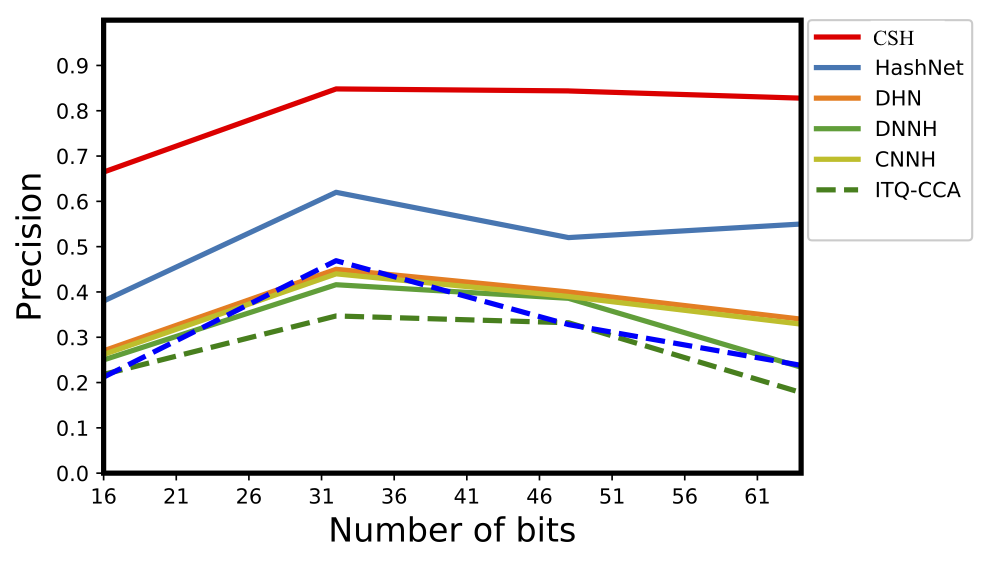}
\label{subfig:image_pr2}
}
\caption{Experimental results of CSQ and compared methods on ImageNet w.r.t. three evaluation metrics.}
\label{fig:imagenet_three_metrics}
\vspace{-15pt}
\end{center}
\end{figure*}

\begin{figure*}[h!]
\begin{center}
\subfigure[P-R curve @64bits]{
\includegraphics[scale=0.32]{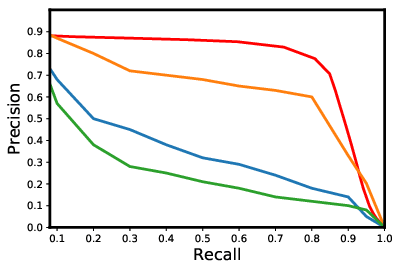}
\label{subfig:video_pr}
}
\subfigure[P@N @64bits]{
\includegraphics[scale=0.32]{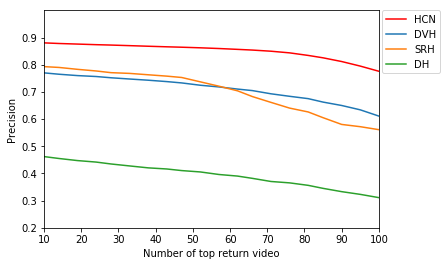}
\label{subfig:video_pn}
}
\subfigure[P@H=2]{
\includegraphics[scale=0.32]{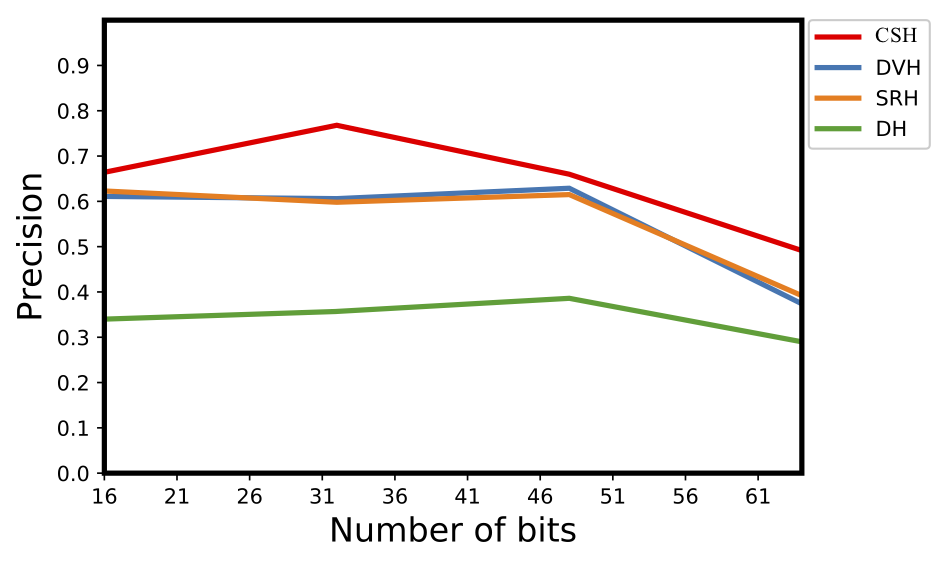}
\label{subfig:video_pr2}
}
\caption{Experimental results of CSQ and compared methods on UCF101 w.r.t. three evaluation metrics.}
\label{fig:ucf101_three_metrics}
\end{center}
\vspace{-20pt}
\end{figure*}

\paragraph{Results} Results in terms of Mean Average Precision (mAP) for image retrieval are given in Tab.~\ref{tab:MAP_image} and~\ref{tab:mAP_2}. In Tab.~\ref{tab:MAP_image}, we take ResNet50 as the backbone for CNNH, DNNH, DHN, HashNet, DCH and our CSQ. In Tab.~\ref{tab:mAP_2}, we take AlexNet and ResNet50 as backbone respectively for five deep methods and our CSQ. From Tab.~\ref{tab:MAP_image}, we can observe that our CSQ achieves the best performance on the image retrieval task. Compared with the state-of-the-art deep hashing methods HashNet and DCH, our CSQ brings an increase of at least 11.5\%, 3.6\%, 3.1\% in mAP for different bits on ImageNet, MS COCO and NUS\_WIDE, respectively. And some retrieval performance boost up to 20\%. Specifically, the mAP boost on ImageNet is much larger than that on the other two datasets, i.e., about 7\%-9\%. Note that ImageNet has the most severe data imbalance among the three image retrieval datasets (Tab.~\ref{tab:datasets}). %This proves that CSQ can efficiently relieve the data imbalance problem. 
From Tab.~\ref{tab:mAP_2}, we can observe that our method achieves superior performance by adopting both AlexNet and ResNet50 as backbone architectures.
%quanhong: check red sentence, right or not
Fig.~\ref{fig:imagenet_three_metrics} shows the retrieval performance in Precision-Recall curves (P-R curve), Precision curves w.r.t.\ different numbers of returned samples (P@N) and Precision curves with Hamming distance 2(P@H=2) respectively on ImageNet. We can find CSQ outperforms all compared methods by large margins on ImageNet w.r.t. the three performance metrics. Additionally, we compare the training time in Tab.~\ref{tab:training_time} and the proposed CSQ achieves a 3 to 5.5 $\times$ faster training speed over DHN and HashNet.
%\vspace{-5pt}

\subsection{Experiments on Video Hashing}

\paragraph{Datasets} Two video retrieval datasets,  UCF101~\cite{soomro2012ucf101} and HMDB51~\cite{kuehne2013hmdb51}, are used with their default settings. On UCF101, we use 9.5k videos for training and retrieval, and 3.8k queries in every split. For HMDB51, we have 3.5k videos for training and retrieval, and 1.5k videos for testing (queries) in each split. 
\vspace{-15pt}
 
\paragraph{Baselines} We compare the retrieval performance of the proposed CSQ against three supervised deep video hashing methods: DH~\cite{qin2017fast}, DLSTM~\cite{zhuang2016dlstm} and SRH~\cite{gu2016supervised} based on the same evaluation metrics as image retrieval experiments.
\vspace{-15pt}

\paragraph{Results} In Tab.~\ref{tab:MAP_video}, our CSQ also achieves significant performance boost on video retrieval. It achieves impressive mAP increases of over 12.0\% and 4.8\% for different bits on UCF101 and HMDB51, respectively. The larger improvements by our method on UCF101 are mainly due to its severe data imbalance. Fig.~\ref{fig:ucf101_three_metrics} shows the retrieval performance in Precision-Recall curves (P-R curve), Precision curves w.r.t.\ different numbers of returned samples (P@N) and Precision curves with Hamming distance 2(P@H=2) respectively on UCF101. From the figure, we can observe that CSQ also outperforms all compared methods by large margins on UCF101 w.r.t. the three performance metrics.  In a nutshell, the proposed CSQ performs consistently well under different evaluation metrics.
\vspace{-5pt}

\begin{table}[h]
\small
\begin{center}
\fontsize{8pt}{12pt}\selectfont
\setlength\tabcolsep{5pt}
\caption{Comparison in mAP of Hamming Ranking for different bits on video retrieval.}
\begin{tabular}{l|c|c|c|c|c|c}
\toprule
\multirow{2}{*}{Method} &\multicolumn{3}{c|}{UCF-101 (mAP@100)} & \multicolumn{3}{c}{HMDB51 (mAP@70)} \\
\cline{2-7}
~ & 16 bits & 32 bits & 64 bits & 16 bits & 32 bits & 64 bits\\
\midrule
{DH}~\cite{qin2017fast} &0.300&0.290&0.470 &0.360&0.360&0.310 \\

{SRH}~\cite{gu2016supervised} &0.716&0.692&0.754 &0.491&0.503&0.509 \\

{DVH}~\cite{liong2017deep} &0.701&0.705&0.712 &0.441&0.456&0.518 \\
\midrule
\textbf{CSQ (Ours)} &\textbf{0.838}&\textbf{0.875}&\textbf{0.874} &\textbf{0.527}&\textbf{0.565}&\textbf{0.579} \\
\bottomrule
\end{tabular}
\label{tab:MAP_video}
\end{center}
\vspace{-27pt}
\end{table}

\paragraph{Differences from image hashing} For video hashing, we need to obtain temporal information by replacing 2D CNN with 3D CNN. In our experiments, CSQ adopts a lightweight 3D CNN, \textit{Multi-Fiber} 3D CNN~\cite{chen2018multi}, as the convolutional layers to learn the features of videos. And the hash layers keep unchanged.
\vspace{-5pt}

\subsection{Visualization}

\begin{figure}[h]
\begin{center}
\setlength{\tabcolsep}{1.5pt}
\includegraphics[scale=0.03]{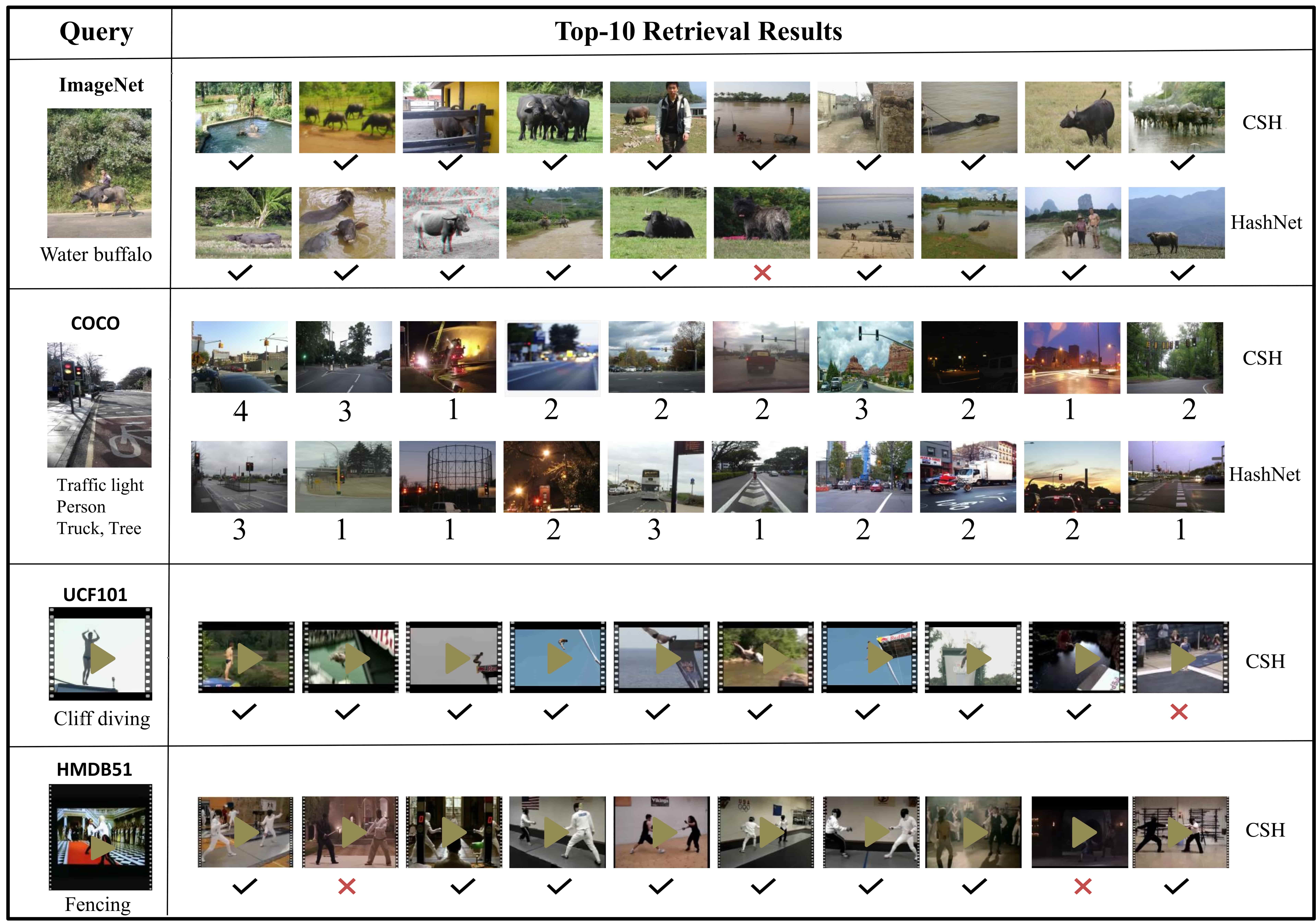}
\caption{Examples of top 10 retrieved images and videos for two image datasets and two video datasets. For COCO images, below each image the number of common labels with the query is given.}
\label{fig:top10}
\vspace{-25pt}
\end{center}
\end{figure}

\paragraph{Visualization of retrieved results} We show the retrieval results on ImageNet, MS COCO, UCF101 and HMDB51 in Fig.~\ref{fig:top10}. It can be seen that CSQ can return much more relevant results. On MS COCO, CSQ uses the centroid  of multiple centers as the hashing target for multi-label data, so the returned images of CSQ share more common labels with the query compared with HashNet.
\vspace{-8pt}

\iffalse
\begin{figure}[h]
\begin{center}
\subfigure[CSQ]{
\includegraphics[scale=0.055]{images/t-sne_a.jpg}
\label{subfig:tsne_CSQ}
}
\subfigure[HashNet]{
\includegraphics[scale=0.055]{images/t-sne_b.jpg}
\label{subfig:tsne_hashnet}
}
\caption{The t-SNE of hash codes learned by proposed CSQ and HashNet. We sample 10k hash codes from random 10 categories.}
\label{fig:t-sne}
\vspace{-15pt}
\end{center}
\end{figure}
\fi

\begin{figure}[t!]
\begin{center}
\setlength{\tabcolsep}{1.5pt}
\begin{tabular}{cc}
\includegraphics[scale=0.03]{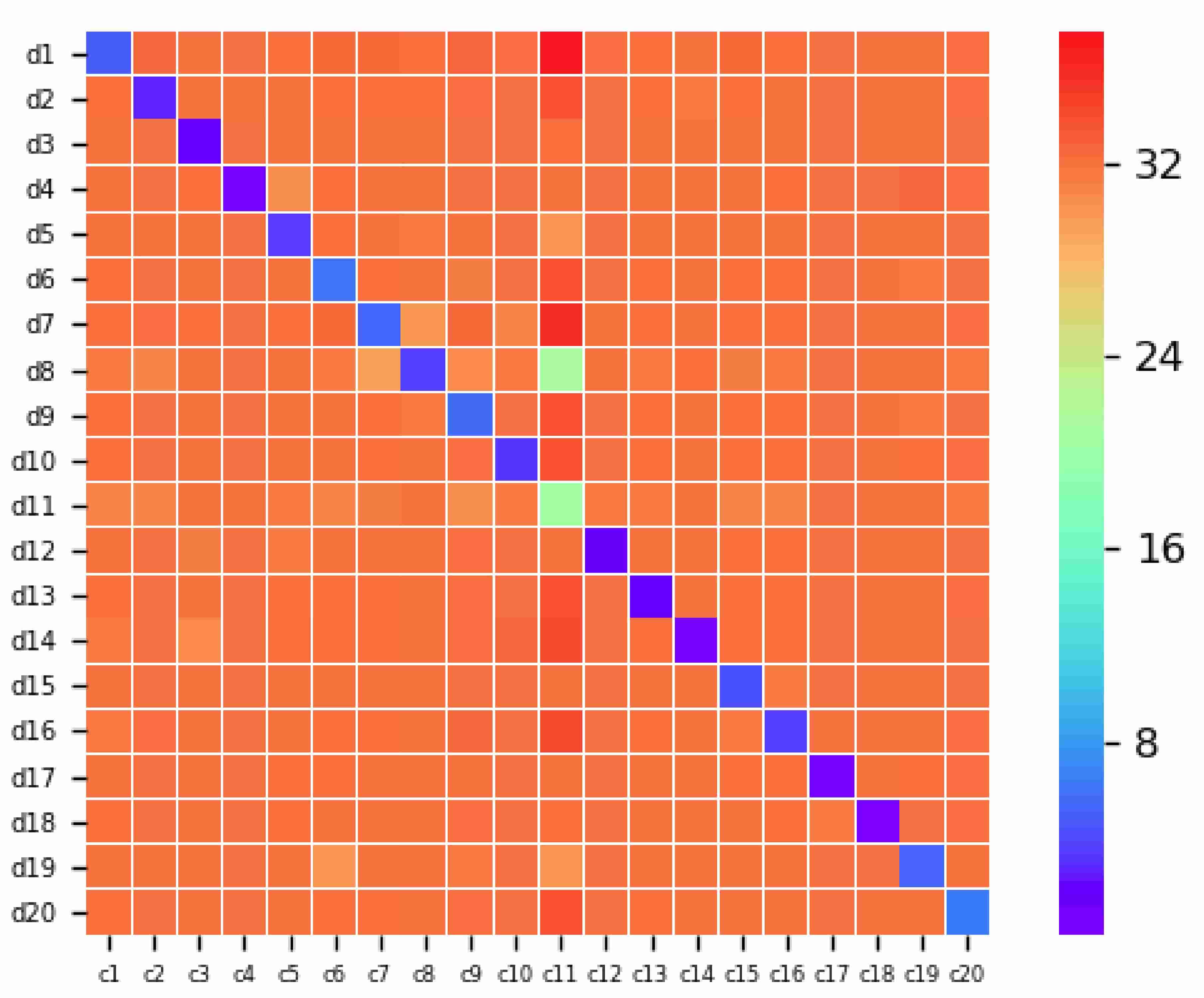}&\
\includegraphics[scale=0.03]{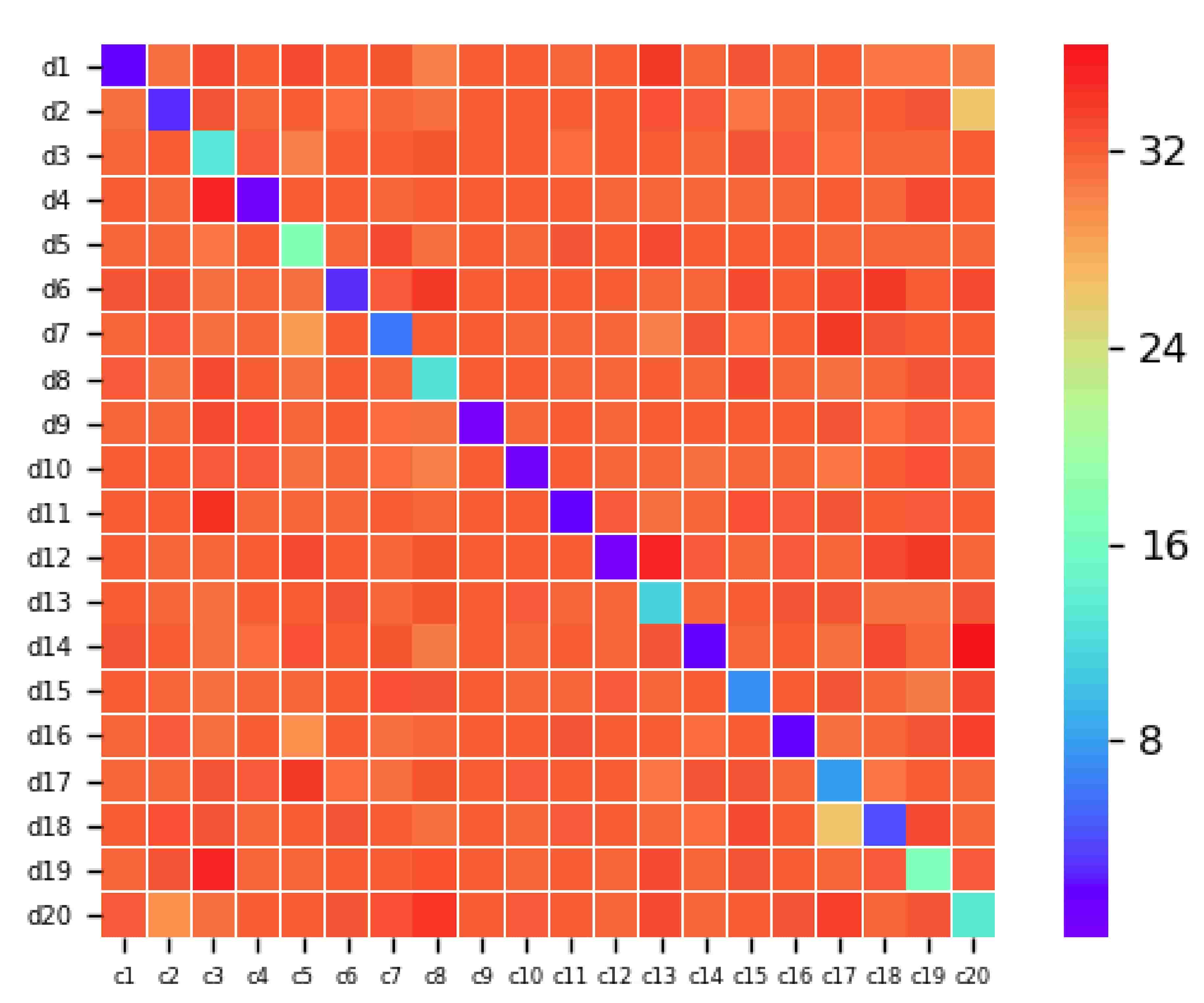}\\
{(a) ImageNet} &{(b) UCF101}
\end{tabular}
\caption{The heat maps of average Hamming distance between 20 hash centers (the columns) with hash codes (64bit, rows) generated by the proposed CSQ from test data in ImageNet and UCF101.}
\label{fig:heat_map}
\vspace{-20pt}
\end{center}
\end{figure}

\vspace{-3mm}
\paragraph{Visualization of hash code distance} We visualize the Hamming distance between 20 hash centers and generated hash codes of ImageNet and UCF101 by heat maps in Fig.~\ref{fig:heat_map}. The columns represent the 20 hash centers of test data in ImageNet (with 1k test images sampled) or UCF101 (with 0.6k test videos sampled). The rows are the generated hash codes assigned to these 20 centers. We calculate the average Hamming distance between hash centers and hash codes assigned to different centers. The diagonal values in the heat maps are the average Hamming distances of the hash codes with the corresponding hash center. We find that the diagonal values are small, meaning the generated hash codes ``collapse'' to the corresponding hash centers in the Hamming space. Most off-diagonal values are very large, meaning that dissimilar data pairs spread sufficiently. %We also find most off-diagonal values are around 32, which is exactly the Hamming distance between different hash centers in a 64 bits space. 

\subsection{Ablation Study}
We investigate the effects of the proposed central similarity, traditional pairwise similarity, and  quantization process for hash function learning, by evaluating different combinations of central similarity loss $L_C$, pairwise similarity loss $L_P$, and quantization loss $L_Q$. The results are summarized  in Tab.~\ref{tab:ablation-study1}. Our CSQ includes $L_C$ and $L_Q$, corresponding to the 1st row in Tab.~\ref{tab:ablation-study1}. When we add $L_P$ to CSQ (2nd row), mAP only increases for some bits. This shows that pairwise similarity has limited effects on further improving over central similarity learning. We add $L_P$ while removing $L_C$ (3rd row), and find that the mAP decreases significantly for various bits. When only using $L_C$, the mAP just decreases slightly. These results show the positive effects of central similarity learning. 

\begin{table}[h]
\small
\begin{center}
\fontsize{8pt}{12pt}\selectfont
\setlength\tabcolsep{4pt}
\caption{The mAP results of CSQ and its three variants on one image dataset and one video dataset.}
\begin{tabular}{lcc|c|c|c|c|c|c}
\toprule
\multicolumn{3}{c|}{}&
\multicolumn{3}{c|}{ImageNet (mAP@1000)} & \multicolumn{3}{c}{UCF101 (mAP@100)} \\
\cline{4-9}
$L_C$ &$L_P$ &$L_Q$ & 16 bits & 32 bits & 64 bits & 16 bits & 32 bits & 64 bits\\
\midrule
\checkmark & &\checkmark &\textbf{0.851}&0.865&\textbf{0.873} &0.838&\textbf{0.875}&0.874 \\

\checkmark &\checkmark &\checkmark &0.847&\textbf{0.870}&0.871 &\textbf{0.840}&0.868&\textbf{0.881} \\

&\checkmark &\checkmark &0.551&0.629&0.655 &0.716&0.739&0.784 \\

\checkmark & &  &0.841&0.864&0.870 &0.824&0.854&0.867 \\
\bottomrule
\end{tabular}
\label{tab:ablation-study1}
\end{center}
\vspace{-25pt}
\end{table}

\subsection{Hash Center Learning}
\label{sec:learn hash center}
In our work, we pre-compute hash centers by using the Hadamard matrix or sampling from a Bernoulli distribution, which is independent of image or video data. We ignore the ``distinctness'' between any two dissimilar data points. For example, the ``distinctness'' between dog and cat should be smaller than that between dog and car. A more intuitive method should be to learn centers from image or video features rather than pre-defined hash centers, which can preserve the similarity information between data points in hash centers. Here we adopt three existing methods to learn centers from features and then compare the learned centers with our pre-computed hash centers, to prove the validity of our proposed hash center generation methods. The three methods to be compared are 1) center learning in face recognition (\textbf{Face Center})~\cite{wen2016discriminative}, 2) center learning in fine-grained classification (\textbf{Magnet Center})~\cite{rippel2015metric}, and 3) center learning in Representative-based Metric Learning (\textbf{RepMet Center})~\cite{karlinsky2019repmet}. We give the details of the three types of learned centers in supplementary material, including loss functions, hyper-parameters and quantization loss to binarize centers for hashing. We apply these learned centers to hashing as the method in Sec~\ref{central similarity quantization} and the retrieval results are given in Tab.~\ref{tab:center_comparison}. We observe the learned centers obtain worse performance than that with our methods. Also, compared with these center learning methods, our pre-computing methods only needs trivial computation/time cost but achieves superior performance. One potential reason is that the fixed hash centers are binary codes while the learned centers are not binary, so we need to do an extra quantization operation on the learned centers, which hurts the similarity information between learned centers and causes worse performances. .
\vspace{-5pt}
\begin{table}[h]
\small
\begin{center}
\fontsize{8pt}{12pt}\selectfont
\setlength\tabcolsep{5pt}
\caption{Comparison between three learned centers with our hash centers on image and video retrieval.}
\begin{tabular}{l|c|c|c|c|c|c}
\toprule
\multirow{2}{*}{Method} & \multicolumn{3}{c|}{ImageNet (mAP@1000)} & \multicolumn{3}{c}{UCF-101 (mAP@100)} \\
\cline{2-7}
~ & 16bits & 32bits & 64bits & 16bits & 32bits & 64bits\\
\midrule
\textbf{Face Center} &0.718&0.723&0.744 &0.693&0.745&0.817 \\

\textbf{Magnet Center} &0.758&0.801&0.818 &0.715&0.793&0.821 \\

\textbf{RepMet Center} &0.804&0.815&0.827 &0.781&0.829&0.835 \\
\midrule
\textbf{Our Center} &\textbf{0.851}&\textbf{0.865}&\textbf{0.873} &\textbf{0.838}&\textbf{0.875}&\textbf{0.874} \\
\bottomrule
\end{tabular}
\label{tab:center_comparison}
\end{center}
\vspace{-25pt}
\end{table}

\section{Conclusion and Future Work}
In this paper, we propose a novel concept ``Hash Center'' to formulate the central similarity for deep hash learning. The proposed Central Similarity Quantization (CSQ) can learn hash codes by optimizing the Hamming distance between hash codes with corresponding hash centers. It is experimentally validated that CSQ can generate high-quality hash codes and yield state-of-the-art performance for both image and video retrieval. In this work, we generate hash centers independently of data  rather than learning from data features, which has been proven effective. In the future, we will continue to explore how to learn better hash centers.
\vspace{-10pt}

\paragraph{Acknowledgement} This work was supported by AI.SG R-263-000-D97-490, NUS ECRA R-263-000-C87-133 and MOE Tier-II R-263-000-D17-112. 

%--------------------------------------

{\small
\bibliographystyle{ieee}
\bibliography{egbib}
}

\end{document}